\documentclass[letterpaper, 10 pt, conference]{ieeeconf}  

\usepackage{graphicx}
\usepackage{amsmath, amsfonts}
\usepackage{booktabs}
\usepackage{tabularx}
\IEEEoverridecommandlockouts                              

\title{\LARGE \bf
Incorporating Explanations into Human-Machine Interfaces \\ for Trust and Situation Awareness in Autonomous Vehicles
}

\author {Shahin Atakishiyev$^{1}$, Mohammad Salameh$^{2}$, Randy Goebel$^{1}$ \thanks{ This work was supported by 
 the Alberta Machine Intelligence Institute (Amii),  Computing Science Department of the University of Alberta, and the Natural Sciences and Engineering Research Council of Canada (NSERC).}
\thanks{Shahin Atakishiyev and Randy Goebel are with the Department of Computing Science, University of Alberta, Edmonton, Alberta, Canada.
}
\thanks{Mohammad Salameh is with Huawei Technologies Canada Co., Ltd., Edmonton, Alberta, Canada.}
\thanks{Correspondence: \tt\small shahin.atakishiyev@ualberta.ca}
}

\begin{document}

\maketitle
\thispagestyle{empty}
\pagestyle{empty}

\begin{abstract}
Autonomous vehicles often make complex decisions via machine learning-based predictive models applied to collected sensor data. While this combination of methods provides a foundation for real-time actions, self-driving behavior primarily remains opaque to end users. In this sense, explainability of real-time decisions is a crucial and natural requirement for building trust in autonomous vehicles. Moreover, as autonomous vehicles still cause serious traffic accidents for various reasons, timely conveyance of upcoming hazards to road users can help improve scene understanding and prevent potential risks. Hence, there is also a need to supply autonomous vehicles with user-friendly interfaces for effective human-machine teaming. Motivated by this problem, we study the role of explainable AI and human-machine interface jointly in building trust in vehicle autonomy. We first present a broad context of the explanatory human-machine systems with the \textit{``3W1H”} (what, whom, when, how) approach. Based on these findings, we present a situation awareness framework for calibrating users' trust in self-driving behavior. Finally, we perform an experiment on our framework, conduct a user study on it, and validate the empirical findings with hypothesis testing.
\end{abstract}

\section{INTRODUCTION}
The emergence of at least partially autonomous driving has shaped a new era in intelligent transportation systems with a promise of safer and eco-friendly roadways. In particular, the DARPA Grand Challenge in 2005 \cite{thrun2006stanley} and further advances in deep neural networks and computer vision algorithms have led to significant breakthroughs in the perception and motion planning ability of autonomous vehicles (AVs) over the last twenty years. While two decades of progress help motivate humans to adopt self-driving cars, traffic accidents and a lack of understanding of the automated decision-making process hinder the widespread acceptance of this technology \cite{atakishiyev2021explainable}. In this sense, \textit{explainable AI (XAI)} has emerged as a crucial component in the design and development of AVs so that a vehicle's decisions are understandable for end users. Another crucial factor is the effective delivery of these explanations so that explanatory descriptions inform safety, transparency, and regulatory compliance perspectives. Hence, this factor also necessitates the integration of  \textit{human-machine interfaces (HMIs)} into highly autonomous driving systems. While XAI is a compendium of approaches to interpreting the action decisions of an autonomous vehicle, HMI is the communication interface that can convey these explanations to road users (human drivers, passengers, pedestrians,
\begin{figure}[t!]
\centering
\includegraphics[width = 1\columnwidth]{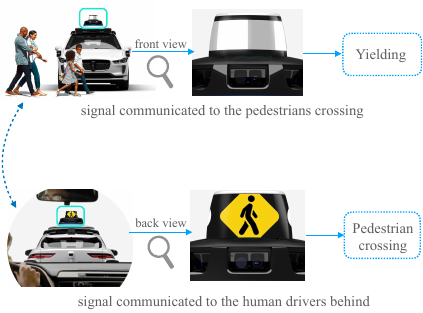}
\caption{An example of explanation communication to the pedestrians (top) and a human driver at the rear (down) by Waymo's self-driving car via its external HMI. The green bounding boxes have been manually added to indicate these signals. The figure drawn based on the content in  \cite{waymo_message_display} and \cite{waymo_driver2023}.}
\label{fig:XAI_HMI}
\end{figure} 
bikers, etc.) during a trip. We provide an example of such interaction in Figure~\ref{fig:XAI_HMI}: Waymo's self-driving vehicle provides explanatory information to the pedestrians in front and the human drivers behind via a simple interface supplied on top of the vehicle. While the pedestrians can interpret this information as ``Waymo is yielding to us and is not going to drive while we cross," the human drivers at the rear may perceive the intent signal in the back view as ``Waymo is stopping because it is yielding to the pedestrians crossing the road." Such a simple explanation provision method shows that HMI is an effective way to describe traffic scenes to road users and help them be aware of the self-driving car's behavioral intention. \\ 
Considering the immense need for AI transparency and the role of human-machine interaction for possible control and self-driving monitoring, our study contributes a structure to the combined role of XAI and HMI for trustworthy autonomous driving. By definition, as XAI aims to provide interpretable models while maintaining good model performance and enabling users to comprehend, trust, and manage the intelligent system, we investigate all these aspects in our study. We specifically analyze the problem setting with a “3W1H” approach: \textit{what} information to deliver, \textit{when} to deliver, \textit{whom} to deliver, and \textit{how} to deliver explanations using a supplied
user interface. In this sense, we systematically

\begin{table*}
  \centering
  \caption{The "3W1H" approach in explanation conveyance to autonomous driving users based on the findings of prior studies}
  \label{tab:3w1h}
  \begin{tabularx}{\linewidth}{>{\raggedright\arraybackslash\hsize=0.22\hsize}X >{\raggedright\arraybackslash\hsize=0.78\hsize}X}
    \toprule
    \multicolumn{1}{l}{\textbf{\hspace{0.5cm}Dimension of explanation}} & \multicolumn{1}{l}{\textbf{\hspace{3cm}Description}} \\
    \midrule
    \hspace{1.5cm}What? & Decisions of an autonomous vehicle, traffic scenes, and events \\
    \addlinespace
    \hspace{1.5cm}Whom? & Passengers, human drivers, people with cognitive and physical impairments, remote operators, bystanders, cyclists, traffic enforcement officials, emergency responders \\
    \addlinespace
    \hspace{1.5cm}When? & Critical and emergent situations, takeover scenarios, the time before an action is performed \\
    \addlinespace
    \hspace{1.5cm}How? & Audio, visual, vibrotactile, text, heads-up display, passenger intervening interface, haptic feedback, braille interface \\
    \bottomrule
  \end{tabularx}
\end{table*}
\hspace{-0.50cm}
review prior investigations, reveal the suggestions, and the practical recommendations derived from that literature. As a result, we incorporate insights from these studies into a framework to improve situation awareness and trust in AVs. \\
Overall, the contributions of our paper are threefold:
\vspace{0.1cm}
\begin{itemize}
    \item We investigate prior studies that explore XAI and HMI for autonomous driving and identify the best practices within the "3W1H" approach;
    \item We present an HMI and XAI-guided situation awareness framework for autonomous driving and perform an experiment on it;
    \item We conduct a user study and validate the experimental findings with human judgment.
\end{itemize} 
\vspace{0.1cm}

\section{Related Work}
The design and use of human interfaces have been of interest to the automotive community since the development of autonomous driving. Human-centric interactive system design does not solely provide driving-related information to users but also can help with meeting their individual needs during a trip with a self-driving vehicle \cite{schieben2019designing,schneider2023don}. In this regard, there are several challenges in the communication of explanations about driving decisions to end users. First, explanations should be relevant to a user's mental model: as user satisfaction is the target focus in the development of autonomous driving systems, the content of explanations should meet their needs \cite{wiegand2020d}. Furthermore, which users (drivers, passengers, external observers) could benefit from these explanations is another aspect of targeting explanations. In addition, the timing mechanism of explanations is also an essential property of the user interface to ensure that the self-driving vehicle's actions or driving scenes are described within appropriate time frames \cite{wiegand2020d}. Finally, how explanations are delivered is also a crucial factor in meeting users' expectations. Considering these four-dimensional properties of explanations in conjunction with HMI, we explore this topic systematically and present the "3W1H" approach. 

\subsection{What?}
What types of explanatory information are needed for road users? Explanations can describe the stationary and dynamic objects in the scene and inform users of how an autonomous vehicle perceives these objects during a journey \cite{kim2023and}. An interactive interface inside a self-driving vehicle can provide updates on traffic situations, weather conditions, and offer customizable preferences (i.e., the temperature inside the vehicle and individual options depending on the users' needs). Moreover, the autonomous car can share information outside to bystanders and other vehicles as a part of the vehicle-to-vehicle (V2V) or vehicle-to-user (V2U) communications.\\
In addition, and as an ongoing research direction, practical XAI methods have recently been explored to describe the rationale behind the decision-making process of AVs in a human-interpretable format \cite{atakishiyev2021explainable}. The critical aspect of explanation techniques is how satisfactory these approaches are from the users' perspective. In this context, users' mental models must be considered: users may favor specific types of explanations depending on the driving scenario. In general, it has been shown that explanations are more related to traffic-related event cognition and describing driving behavior of AVs as an answer to the ``What" question in this context \cite{schneider2023don}.   
\subsection{When?}
While a history of continuous action- and scene-explanation pairs can be recorded for possible post-trip driving analysis and forensic investigations, it is noteworthy to mention that users, particularly passengers and human drivers do not always need explanations. According to Koo et al.'s study \cite{koo2015did}, seamlessly delivered explanations can result in \textit{mental overload} for human drivers in semi-autonomous driving and distract them from meeting their primary obligations. Instead, they can be alerted when an autonomous mode decision is initiated, when driving conditions change substantially, when takeover requests are made, and when sensor failure occurs. In another user study, Haspiel et al. \cite{haspiel2018explanations} have focused on the significance of timing for explanations, and their exploration concludes that human drivers favor explanations just \textit{before} action is decided rather than receiving explanations
 \textit{after} the action is performed. Avoiding a potentially overwhelming information flow also applies to explanation provision for passengers. Kim et al.s' \cite{kim2023and} recent user study on a real road with a wizard experimenter reveals that users favor explanations in critical and risky traffic circumstances rather than having continuously presented behavioral information. Similarly, Shen et al. \cite{shen2022to} have also qualitatively validated the premise that people favor explanations in near-crash and emergent situations. 

 \subsection{Whom?}
User categories encompass individuals who directly interact with a self-driving vehicle (i.e., passengers and backup drivers) or are affected by its presence and operation on roads \cite{atakishiyev2021explainable}. Explanatory behavior can be conveyed to road users outside the vehicle (i.e., bystanders, pedestrians, and cyclists) using language or intent signals as well as shown in Figure \ref{fig:XAI_HMI} to ensure that individuals nearby remain aware of the behavior of the autonomous vehicle while it operates. Moreover, people outside and adjacent to an autonomous vehicle, such as bystanders, pedestrians at crosswalks, drivers operating non-autonomous cars, and cyclists, may also expect behavioral information from an autonomous car. Additionally, understanding the decisions of a self-driving vehicle may be a necessity for traffic enforcement officials \cite{hansson2021self} and emergency responders \cite{liu2023first}. To ensure operational safety, traffic enforcement personnel may carry out a compliance check on the self-driving vehicle at some time and the vehicle may be requested to provide some form of explanation on its motion. Furthermore, behavioral information may be helpful for emergency responders for effective response to the accident and emergency conditions caused by/with the presence of an autonomous vehicle. Overall, explanations should be conveyed to relevant interaction partners while an autonomous car operates on roads.

\subsection{How?}
Various studies have shown that users might have different preferences for receiving explanations depending on their identity and traffic situation. For example, Faltaous et al.s' \cite{faltaous2018design} user study in a simulation environment shows that providing multimodal explanations with \textit{auditory, visual}, and \textit{vibrotactile} feedback is effective for potential takeover requests in highly-urgent driving conditions. Furthermore, Schneider et al. \cite{schneider2021increasing} have tested five various feedback techniques - \textit{light, audio, object visualization, textual information}, and \textit{vibration} - in a virtual driving scene, and evaluate user satisfaction with these explanation modalities. Their findings show that light or object visualization is preferred more for proactive situations, while sound and light are more favored for reactive scenarios. Interestingly, the users are not satisfied with textual descriptions or sound-based feedback and even find the latter disturbing in the long term. Detjen et al. \cite{detjen2021towards} show that a \textit{planar heads-up display} (pHUD) and \textit{contact analog heads-up display} (cHUD), as an augmented reality presenter, have been liked by testers as an effective display technique in both low and highly automated driving settings. In contrast, Dandekar et al. \cite{dandekar2022display} have investigated an effective way of conveying driving information to passengers while not distracting them from their in-vehicle activities. The users' feedback indicates that a colored light bar display and windshield display are satisfactory interfaces and do not disturb them while being immersed in non-driving related infotainment. Kim et al. \cite{kim2023and} also conclude that the semantic segmentation map of objects displayed via windshield display instills trust in human participants in the behavior of an autonomous vehicle. \\
It is also noteworthy to state that automotive HMI design should consider humans' varying physical and cognitive capabilities and be inclusive of everyone. Particularly, people with some form of physical and cognitive impairment may need customized user interfaces. In recent research, Arfini et al. \cite{arfini2023design} have presented theoretical and pragmatic challenges in the effective interface design for people with some functional limitations. Autonomous vehicle users may have visual, hearing, mobility, and speech impairments, and customized user interfaces may help them get explanatory information on the vehicle's actions and traffic situations. For instance, a braille interface can help people with visual difficulties to acquire driving-related information from the vehicle. Or a gesture recognition system may be implemented to track the user’s hand movements and enable them to interact with the vehicular interface. Hence, in general, we can summarize that no single type of HMI can meet the needs of all autonomous car users due to their diverse cognitive and physical abilities and personal preferences. \\
While the "3W1H" approach reveals a broad spectrum of explanation conveyance to end users  (see Table \ref{tab:3w1h}), the goal remains the same: bringing situation awareness to people inside and outside of an autonomous car. In the next subsection, we present a unified approach to a general situation awareness framework via XAI and HMI and describe the implications of such a framework for people in the loop.  

\section{A Unified Approach to Situation Awareness via XAI and HMI}
The goal of an XAI system is to provide explanatory information on the system’s particular decisions or behavior. According to the definition of the term, \textit{situation awareness} by Endsley \cite{endsley1995measurement}, users in the loop must be aware of what particular decision the AI system made,  why it made that specific decision, and what decisions it will make in the next similar state at a later time. The three concepts have been referred to as \textit{perception}, \textit{comprehension}, and \textit{projection}, and are viewed as indicators of situation awareness in human-in- 

\begin{figure*}[htp!]
    \centering
    \includegraphics[width=17cm]{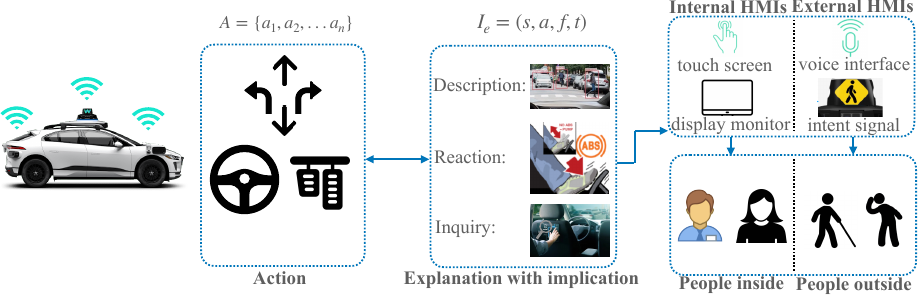}
    \caption{The proposed situation awareness framework for inside and outside users of an autonomous vehicle with XAI and representative HMIs}
    \label{fig:SA}
\end{figure*}
\hspace{-0.48cm}
the-loop AI systems. Sanneman and Shah \cite{sanneman2022situation} have further expanded that framework from an informativeness perspective. They correspond the aforementioned concepts into three levels of explanations - Level 1, Level 2, and Level 3 - for human-centered XAI systems, respectively. 
Here we adapt Sanneman and Shah's \cite{sanneman2022situation} framework to autonomous driving scenarios and describe the implications of such a framework for self-driving vehicle users. \\
Given a traffic scenario $s \in S$, we denote a set of possible actions (i.e., go straight, turn left, etc.) that may be taken by an autonomous car as \[A = \{a_1, a_2, ..., a_n\}.\] At each time step of motion of the car, an explanation interface may describe what action the car is performing and why it is doing so. We can describe this particular action as  $a \in A$ and the causal factor (i.e., traffic light, pedestrians crossing, etc.) that made the car behave in that particular way as  $f \in F$. The explanatory information $I_e$ for situation awareness can then be described via a combination of \textit{four elements}: the traffic scenario, chosen action, causal factor inducing that action, and explanation delivery time: \begin{equation}
I_e = (s, a, f, t)
\end{equation}
This explanation is communicated to autonomous vehicle users using various built-in internal and external HMIs as needed. Depending on the interplay between the user and the explanation, the information conveyed can be classified as interactive or non-interactive, which can be stated as follows: 

\[
I_e=
\begin{cases}
    \textit{interactive}, & \text{if reactive or inquisitive }   \\
    \textit{non-interactive}, & \text{if descriptive}
\end{cases}
\]
Now, let us define what we mean by \textit{descriptive, reactive}, and \textit{inquisitive explanations} in this context: \\
 \textbf{1. Descriptive explanations}: A user interface provides general information about the action of the self-driving car and the scene it is moving through. The people inside or outside become aware of the driving environment and behavior of the self-driving vehicle without further interference. \\
\textbf{2. Reactive explanations}: The user interface invites the passenger or human driver to react to an emerging situation. For passengers, this could be an emergency override scenario where vehicular automation by design allows them to take control of vehicles in case other outreach actors can not control the vehicle. For human drivers, these are typically takeover requests, where the explanation interface communicates upcoming emergent states, and the driver must take control of the vehicle from the automated mode and manage the situation safely. \\
\textbf{3. Inquisitive explanations}: While descriptive and reactive explanations are already well-known to the automotive community, to the best of our knowledge, inquisitive explanations have not been investigated well in the current literature. Inquisitive explanations refer to the explanations that users, such as passengers, can ask the system to follow up on the previous response or to test the robustness of the interactive user interface. These questions can be anything related to the autonomous vehicle's decisions or traffic scene in general, but can also be ``tricky'' or \textit{adversarial} questions aiming to stress test the explanation interface. For instance, assume in an actual \textit{right turn} scenario under a green light, a passenger asks the conversational user interface, "Why is the car turning to the \textit{left?}" as an adversarial question. In this case, the user interface should provide an action-reflecting response like ``No, the car is turning to the right as the traffic light allows a right turn,'' explaining what the autonomous car is doing and why it is doing so. These questions may be asked deliberately or unintentionally (e.g., people with visual impairments may have difficulty perceiving the scene correctly) to test the trustworthiness of the vehicle, its actions, and its awareness of the surroundings. As a consequence, a human-machine interface should not only provide conventional explanations but also defend against potential adversarial queries to ensure that explanations are scene and action-reflecting. This feature is an essential property of automotive HMI. The ability to provide correct explanations can instill trust in users and encourage them to continue to use AVs. The graphical description of this framework is shown in Figure \ref{fig:SA}. 
With this approach, the proposed situation awareness framework may achieve Level 1 (chosen action) and Level 2 explanations (explanation for the chosen action), i.e., perception and comprehension, as described by Sanneman and Shah \cite{sanneman2022situation}. These authors describe Level 3 explanations as ``XAI for projection'' and relate it to counterfactual analysis, such as what an autonomous car will do if the driving scene changes in a certain way. Perhaps in the context of autonomous driving, Level 1 and Level 2 explanations are more essential for the users as they are more concerned about the actual scene the vehicle is getting through. With that said, curiosity to assess automation ability and the reaction of the vehicle to further emergent driving conditions may necessitate Level 3 explanations for foreseeable situation awareness as well.

\begin{figure*}[htp!]
    \centering
    \includegraphics[width=17.6cm]{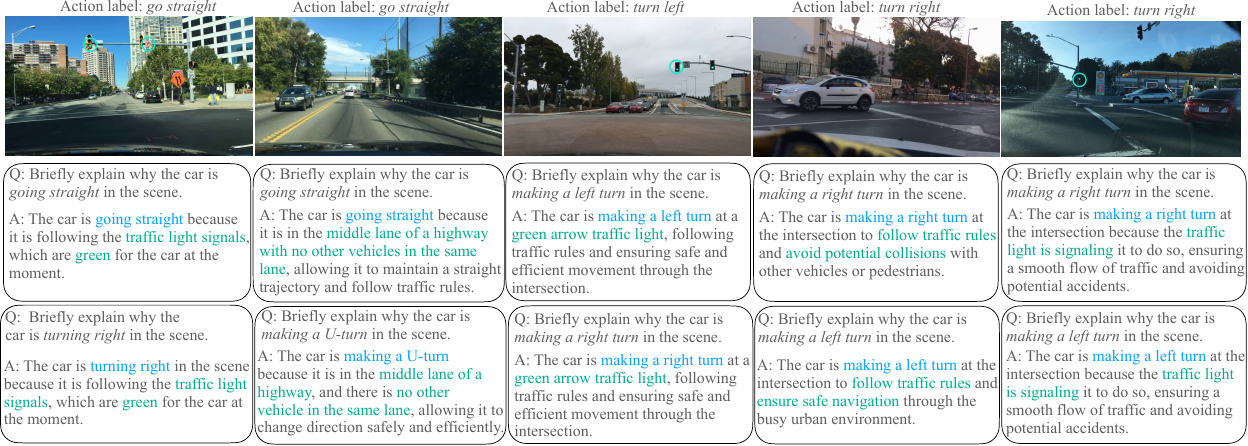}
    \caption{Our experiment on the five chosen traffic scenes from the BDD-A dataset with the LLaVA multimodal transformer. While LLaVA seems to yield correct explanations on \textit{conventional questions} (top) with actual actions (blue-colored text) + causal factors (green-colored text), it fails to generate factual explanations on the \textit{adversarial} questions (bottom). The bounding boxes have manually been added to indicate causal factors inducing the chosen actions.}
    \label{fig:LLaVA_exp}
\end{figure*}

\section{Case Study: Interactive Dialogues between a User and An Autonomous Vehicle}
To validate our proposed framework, we perform an initial and simple empirical study and use human judgment to evaluate the effectiveness of the framework on the chosen traffic scenarios. As HMI can be both interactive and descriptive from an explanation communication perspective, we choose to design interactive explanations so that a user can prompt their question and expect context-aware information from an explanation interface. We formalize this setting as a  \textit{visual question answering} (VQA) problem. VQA is a learning task at the intersection of computer vision and natural language processing that inputs an image \textit{i}, a textual question \textit{q} associated with the content of the image, and predicts an answer \textit{a} for the asked question. The problem can be more precisely represented as a tuple of the specified parameters:  \[x= (i, q, a)\]
Most of the state-of-the-art VQA models input an image and question parameters, then output a combined representation of vision and language $ r \in \mathbb{R}^{d_r}$ via a multimodal learning network \textit{f}: 

\[r= f(i,q)\]
Asking a question about the behavior of a vehicle is driven, at least partly, by human intuition. Becoming aware of traffic events and a human driver's decisions in the car helps passengers on board understand ongoing situations and have a comfortable trip. This intuition also can be related to asking meaningful questions to the conversational HMI of an autonomous vehicle about the driving scenes. 
We have recently performed a preliminary experimental study on this concept and showed its applicability to explainable autonomous driving \cite{atakishiyev2023exp}. A crucial aspect of such interaction is the delivery of context-aware and action-reflecting explanations in a timely manner. In this sense, we design our experiment as follows: We use Large Language and Vision Assistant, LLaVA \cite{liu2024visual}, as a multimodal learning framework that inputs vision and textual query for describing driving scenes. LLaVA is a multimodal transformer architecture built on top of the Vicuna LLM \cite{vicuna2023} and the pre-trained CLIP visual encoder ViT-L/14 \cite{radford2021learning}. We sample driving scenes from the real driving videos of the Berkeley DeepDrive Attention (BDD-A) dataset \cite{xia2019predicting}, and select five different scenarios shown in Figure \ref{fig:LLaVA_exp}. We then consider two types - \textit{conventional} and \textit{adversarial} - questions for the actions performed in these scenes. While the former is context-aware and action-reflecting, the adversarial question refers to asking an \textit{incorrect} questions about actions, deliberately or unintentionally,  to stress test the robustness of the explanation provision model. The key idea with this test is that HMI must always communicate correct explanations and defend against human adversarial questions by providing context-aware information.\\
We present the explanations (i.e., responses) generated by LLaVA on the action-reflecting and adversarial questions in Figure \ref{fig:LLaVA_exp}. As seen, the model generates reasonable and human-interpretable explanations for conventional questions and justifies its responses. However, a tricky question easily confuses the model and makes it generate an incorrect answer to the asked question. For instance, in the leftmost scene in Figure \ref{fig:LLaVA_exp}, the car moves straight because of the green light (i.e., inside the left bounding box). Furthermore, there is a sign (i.e., inside the right bounding box) that \textit{prohibits} the car from \textit{making a right turn} at that intersection. Ideally, when we ask that adversarial question about the right turn, we might be satisfied with an explanation like ``No, the car is going straight and no right turn is allowed in the scene as the traffic signal prohibits it." However, we observe that the model was flawed with such a deliberate question and failed to present an adequate answer. The higher influence of a textual question on a VQA model's prediction is generally known as the \textit{language prior problem}. Therefore, the possibility of adversarial questions must be considered in the construction mechanism of automotive VQA models, and relevant HMIs must provide corrective and context-aware information against such questions.
\begin{figure*}[htp!]
    \centering
    \includegraphics[width=17.6cm]{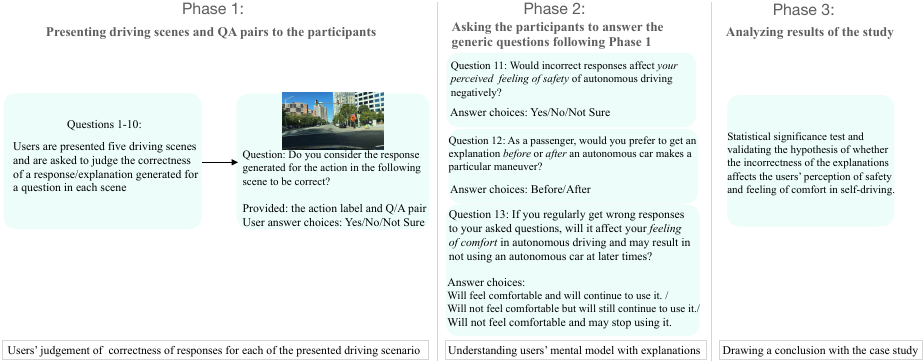}
    \caption{Design of the case study based on the experiment in Figure \ref{fig:LLaVA_exp}. The participants judge the correctness of explanations for each of the five scenes presented. After getting experience with explanations, they are asked two more questions on their perceived safety and mental comfort with the role of explanations while using an autonomous vehicle. Users' responses are validated with a statistical significance test to draw a conclusion with the case study.}
    \label{fig:user_study}
\end{figure*}
\hspace{-0.50cm}\\
We need to consider that an autonomous vehicle is equipped with an interactive HMI so that people inside the vehicle can ask such questions at some point in the trip with the vehicle. To understand the impact of the generated responses on the users' feeling of comfort and safety perception in autonomous driving, we conduct a human study with the above experimental framework, considering them as passengers of an autonomous vehicle.\\

\subsection{Design of the user study} We have performed a user study\footnote{The user study was conducted under the Research Ethics principles of the University of Alberta, and all the participants were paid at an equal rate.} (see Figure \ref{fig:user_study}) with 20 participants (10 males, 10 females) with an age range from 20 to 47 (average=28.65, standard deviation=6.03). The participants are university students and employees in an industrial sector with diverse technical backgrounds who drive or use public transportation regularly in everyday life. We provide participants with an online questionnaire on the QA pairs described in five scenes in Figure \ref{fig:LLaVA_exp} in the first step. Once the participants finish the assessment of the generated responses for the questions on the autonomous car's actions, they are further asked three generic questions: their perceived safety, feeling of comfort with incorrect explanations, and timing preference. Overall, each respondent answers thirteen questions in their 20- to 25-minute study engagement.\\
\subsection{Analysis of the results} \textit{Phase 1:} While analyzing the respondents’ judgment of the explanation correctness, we see that most of them find explanations generated for conventional questions satisfactory (see Table \ref{table:phase_1}). While Scenario 1 and Scenario 3 explanations are assessed as correct explanations by 100\% and 95\% of the users, they are skeptical about some explanations in Scenario 2, 4, and 5, leading to 75\%, 60\%, and 60\%, respectively. On the other hand, the participants are more confident in judging the responses for the adversarial questions, and 100\% of them spot flawed explanations for that type of question in Scenarios 1, 2, and 5. Just one and two of them are unsure about the incorrectness of the responses to the adversarial questions in Scenario 3 and 4, respectively. It turns out that detecting an incorrect answer to a question is easier for humans, likely as a consequence of reasoning; While getting irrational responses to a query, even at the beginning of a response, easily triggers a response of “No.”  Note that the users are usually firmer in judging every detail of a rationale to say “Yes,” and autonomous driving explanations by such a judgment criterion are not an exception. \\
\textit{Phase 2:}
After the participants complete Phase 1 and understand the role of explanations in autonomous driving, we evaluate 1) their preference for time to deliver an explanation and 2) how the faithfulness of explanations affects their mental model in terms of their perceived feeling of safety and comfort in autonomous driving. 100\% of the participants have preferred to get an explanation \textit{before} an autonomous vehicle makes a particular maneuver rather than having it \textit{after}. Such a preference is reasonable as prior explanations elucidate what an autonomous car is going to do next and help users become aware of the situation and monitor the safety of the subsequent action. \\ 
To evaluate the impact of incorrect explanations on the users' feeling of comfort and perceived safety in autonomous driving, we perform simple hypothesis testing. As the minimum percentage of ``majority voting'' in the users' judgment of explanation correctness is 60\% (see Table \ref{table:phase_1}), we use this number in the validation of the users' safety perception and feeling of comfort in our hypotheses. More specifically, we define the following null and alternative hypotheses:
\begin{figure*}[htp!]
    \centering
    \includegraphics[width=17.6cm]{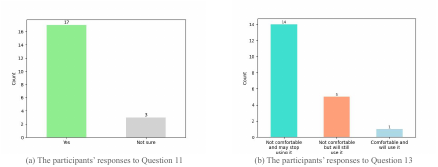}
    \caption{The participants' responses to Question 11 and Question 13 in Fig \ref{fig:user_study} on their perceived feeling of safety and comfort with incorrect explanations}
    \label{fig:survey_res}
\end{figure*} \\
\textit{Null hypothesis} on implication in perceived safety: Incorrectness of explanations will affect 60\% of the respondents negatively on their safety perception in autonomous driving. \\
\textit{Alternative hypothesis (right-tailed)}: Incorrectness of explanations will affect \textit{more than} 60\% of the respondents negatively on their safety perception in autonomous driving. \\
We use the one-proportion Z-test method to validate our hypotheses. The one-proportion Z-test is defined as follows:
\begin{equation}
Z = \frac{\hat{p} - p_0}{\sqrt{\frac{p_0(1-p_0)}{n}}}
\end{equation}
where $\hat{p}$ is the sample proportion, $p_0$ is hypothesized population size, and $n$ is the sample size. Given $\hat{p}$=\(\frac{17}{20}\)=0.85, $p_0$=0.6, and n=20, putting these numbers into the formula gives Z$\approx$2.28. We consider a significance level of $\alpha$=0.05. For the right-tailed hypothesis, the critical Z-value is 1.645 \cite{prob_tables}, and our Z-value of $\approx$2.28 is \textit{greater} than the critical Z-score. So, we \textit{reject} the null hypothesis and conclude that incorrectness of explanations affects \textit{more than} 60\% of the participants negatively on their perceived feeling of safety. \\
Similarly, we define the following null and alternative hypotheses for implications of incorrect explanations on the participants' feeling of comfort with an autonomous vehicle:\\
\textit{Null hypothesis} on implication in the feeling of comfort: Incorrectness of explanations will affect 60\% of the respondents negatively on their feeling of comfort. \\
\textit{Alternative hypothesis (right-tailed)}: Incorrectness of explanations will affect \textit{more than} 60\% of the respondents negatively on their feeling of comfort.\\
Referring to Figure \ref{fig:survey_res}(b), $\hat{p}$=\(\frac{14}{20}\)=0.7, $p_0$=0.6, n=20, and a significance level of $\alpha$=0.05, we get a Z-score of $\approx$0.91, which is \textit{less} than the critical Z-value, 1.645. Hence, we \textit{fail to reject} the null hypothesis as there is not enough evidence to suggest that the proportion of the participants who think incorrect explanations affect their feeling of comfort in an
autonomous vehicle negatively is greater than 60\%. 
\begin{table}[!t]
  \centering
  \caption{The participants' judgment of the correctness of explanations on the conventional and adversarial question pairs for each scenario described in Figure \ref{fig:LLaVA_exp}.}
  \label{table:phase_1}
  \small
  \begin{tabularx}{0.48\textwidth}{>{\centering\arraybackslash}p{0.07\textwidth}>{\centering\arraybackslash}p{0.17\textwidth}>{\centering\arraybackslash}p{0.17\textwidth}}
    \hline
    \textbf{Driving scenario (left to right)} & \textbf{ Distribution of the participants' answers to the conventional questions (Yes/No/Not sure)} & \textbf{Distribution of the participants' answers to the adversarial questions (Yes/No/Not sure)} \\
    \hline
    \#1 & 20/0/0 & 0/20/0 \\
    \hline
    \#2 & 15/3/2 & 0/20/0 \\
    \hline
    \#3 & 19/0/1 & 0/19/1 \\
    \hline
    \#4 & 12/2/6 & 0/18/2 \\
    \hline
    \#5 & 12/4/4 & 0/20/0 \\
    \hline
  \end{tabularx}
\end{table} 

\subsection{Limitations}
Our study is relatively small and has several limitations. First, we conduct the user study via an online questionnaire, and we do not know how the participants' answers may change in case these questions are presented in simulated augmented reality or real AVs. In addition, while we underscore the essence of considering people's various cognitive and functional abilities for acquiring relevant messages from HMIs, we have not addressed this nuance in our experiment and user study. Moreover, there is a well-known Autonomous Vehicle Acceptance Model (AVAM) proposed by Hewitt et al. \cite{hewitt2019assessing} that assesses user acceptance of AVs with nine essential factors. However, interestingly, that model does not consider explainability as one of those key factors. So, we argue that the AVAM model enhanced with the explainability criterion would be a stronger acceptance model for AVs. Finally, a larger-scale user study with different groups and more diverse scenarios can help us draw stronger conclusions with this work in the next phase of our research.
\subsection{Summary of the findings}
The main outcomes of our study are summarized as follows: \\
(1) \textit{Timing matters}: Time to deliver an explanation is an essential feature for autonomous driving users. Particularly, delivering prior and time-sensitive explanations helps them understand the driving behavior of an autonomous vehicle and be aware of the vehicle's subsequent intention.  \\
(2) \textit{Robustness matters}: Our experiment shows that even advanced explanation models may fail to provide adequate responses to human adversarial questions.  Flawed explanations may have a negative impact on the users' feeling of safety and trust in the automation ability of a self-driving car. Consequently, explanation interfaces must always understand users' conventional and deliberate questions, defend against adversarial queries, and provide faithful explanations for effective and reliable human-machine interaction. \\
(3) \textit{Inclusivity matters}: Automotive HMIs should not only take people's technical backgrounds into account but also consider their various functional and cognitive capabilities for communicating explanations to them. Hence, need-based and customized HMIs must be a key aspect in fostering everyone-inclusive autonomous driving and meeting general society's expectations of this technology.     

\section{Conclusion}
We have presented a situation awareness framework for autonomous driving backed by HMI and explanations. Our experiment and humans' judgment of experimental findings shows that faithfulness in explanations is of paramount importance for users to understand driving situations, trust action decisions, and the safety of self-driving vehicles. We believe that our work can help enhance the transparency and safety of this technology, inform effective automotive HMI design, and promote everyone-inclusive autonomous driving. 
\bibliography{IEEEabrv.bib, main.bib}{}
\bibliographystyle{IEEEtran}

\end{document}